\documentclass[a4paper,fleqn]{cas-sc}

\usepackage[authoryear]{natbib} % 删除了 longnamesfirst，使引用更简洁
\usepackage{subcaption} 
\usepackage[ruled,vlined,linesnumbered]{algorithm2e}
\usepackage{tabularx}
\usepackage{calc}

\def\tsc#1{\csdef{#1}{\textsc{\lowercase{#1}}\xspace}}
\tsc{WGM}
\tsc{QE}

\begin{document}
\let\WriteBookmarks\relax
\def\floatpagepagefraction{1}
\def\textpagefraction{.001}

% --- 1. 页面页眉信息 ---
\shorttitle{SIGMAE: A Spectral-Index-Guided Foundation Model for Multispectral Remote Sensing}
\shortauthors{X. Zhang et al.}

% --- 2. 论文标题 (去掉了基金标记 tnotemark) ---
\title [mode = title]{SIGMAE: A Spectral-Index-Guided Foundation Model for Multispectral Remote Sensing}

% --- 3. 作者列表 ---

% 第一作者 & 通讯作者
\author[1]{Xiaokang Zhang}[orcid=0000-0002-6127-4801]
% \cormark[1] 
\ead{zhangxiaokang@whu.edu.cn} 
\credit{Conceptualization, Methodology, Writing - original draft, Funding acquisition}

\author[2]{Bo Li}
\credit{Data curation, Investigation, Software, Validation, Writing - review \& editing}

\author[3]{Chufeng Zhou}
\credit{Data curation, Investigation, Software, Validation, Writing - review \& editing}

\author[4]{Weikang Yu}
\credit{Data curation, Investigation, Software, Validation, Writing - review \& editing}

\author[5]{Lefei Zhang}
\credit{Conceptualization, Methodology, Supervision, Writing - review \& editing}

% --- 4. 单位地址 ---
\affiliation[1]{organization={School of Artificial Intelligence, Wuhan University},
            city={Wuhan},
            postcode={430072}, 
            country={China}}
            
\affiliation[2]{organization={School of Artificial Intelligence and Automation, Wuhan University of Science and Technology},
            city={Wuhan},
            postcode={430081}, 
            country={China}}

\affiliation[3]{organization={School of Electronic Information, Wuhan University of Science and Technology},
            city={Wuhan},
            postcode={430081}, 
            country={China}}

\affiliation[4]{organization={Helmholtz-Zentrum Dresden-Rossendorf},
            city={Freiberg},
            postcode={09599}, 
            country={Germany}}

\affiliation[5]{organization={School of Computer Science, Wuhan University},
            city={Wuhan},
            postcode={430072}, 
            country={China}}

% --- 5. 通讯作者脚注文本 (对应样刊第一页左下角效果) ---
% \cortext[1]{Corresponding author}

% --- 6. 摘要 (Abstract) ---
\begin{abstract}
Pretraining and fine-tuning have emerged as a new paradigm in remote sensing image interpretation. Among them, Masked Autoencoder (MAE)-based pretraining stands out for its strong capability to learn general feature representations via reconstructing masked image regions. However, applying MAE to multispectral remote sensing images remains challenging due to complex backgrounds, indistinct targets, and the lack of semantic guidance during masking, which hinders the learning of underlying structures and meaningful spatial-spectral features. To address this, we propose a simple yet effective approach, Spectral Index-Guided MAE (SIGMAE), for multispectral image pretraining. The core idea is to incorporate domain-specific spectral indices as prior knowledge to guide dynamic token masking toward informative regions. SIGMAE introduces Semantic Saliency-Guided Dynamic Token Masking (SSDTM), a curriculum-style strategy that quantifies each patch’s semantic richness and internal heterogeneity to adaptively select the most informative tokens during training. By prioritizing semantically salient regions and progressively increasing sample difficulty, SSDTM enhances spectrally rich and structurally aware representation learning, mitigates overfitting, and reduces redundant computation compared with random masking. Extensive experiments on five widely used datasets covering various downstream tasks, including scene classification, semantic segmentation, object extraction and change detection, demonstrate that SIGMAE outperforms other pretrained geospatial foundation models. Moreover, it exhibits strong spatial-spectral reconstruction capability, even with a 90\% mask ratio, and improves complex target recognition under limited labeled data. The source codes and model weights will be released at https://github.com/zxk688/SIGMAE.
\end{abstract}

% --- 7. 关键词 ---

% \begin{highlights}
% \item A novel spectral-index-guided MAE (SIGMAE) is proposed for multispectral image pretraining.
% \item Semantic Saliency-guided Dynamic Token Masking (SSDTM) is proposed to integrate spectral priors and curriculum learning for efficient representation learning.
% \item The developed foundation model achieves SOTA performance on various geospatial tasks with high efficiency.
% \end{highlights}

\maketitle

% --- 正文开始 ---

\section{Introduction}
Multispectral satellite images have been widely used for large-scale and continuous Earth observation applications like land use/cover mapping, natural resource management, environmental monitoring, and disaster assessment \citep{benhammou2022sentinel2globallulc,wang2023ssl4eo,sun2024segment,ghorbanzadeh2022outcome,zhang2023cross}.
A fundamental requirement for these applications is to efficiently and accurately capture desired information through image recognition and interpretation techniques.
With the rise of deep learning, numerous models based on convolutional neural networks (CNNs) and Transformers have been proposed to learn discriminative feature representations \citep{zhang2016deep,9244062}. 
Despite these efforts, these approaches still rely on large amounts of labeled data, and fully exploiting the rich spectral and spatial information in multispectral imagery remains
a critical challenge  \citep{tong2020land,zhang2022artificial}.

The rapid advancement of self-supervised learning has driven the emergence of foundation models, establishing a new paradigm for geospatial applications \citep{xiao2025foundation,lu2025vision}.
By deriving supervisory signals directly from large-scale unlabeled data, pretraining strategies—such as contrastive learning \citep{li2022global,guo2024skysense} and masked image modeling \citep{prithvi,mendieta2023towards,noman2024rethinking}—alleviate the need for extensive human annotations.
The pretrained models can subsequently be fine-tuned with only a limited number of labeled samples, significantly enhancing efficiency and adaptability in downstream Earth observation tasks \citep{ghamisi2025geospatial,11215798}.

\begin{figure}
    \centering
\includegraphics[width=\linewidth]{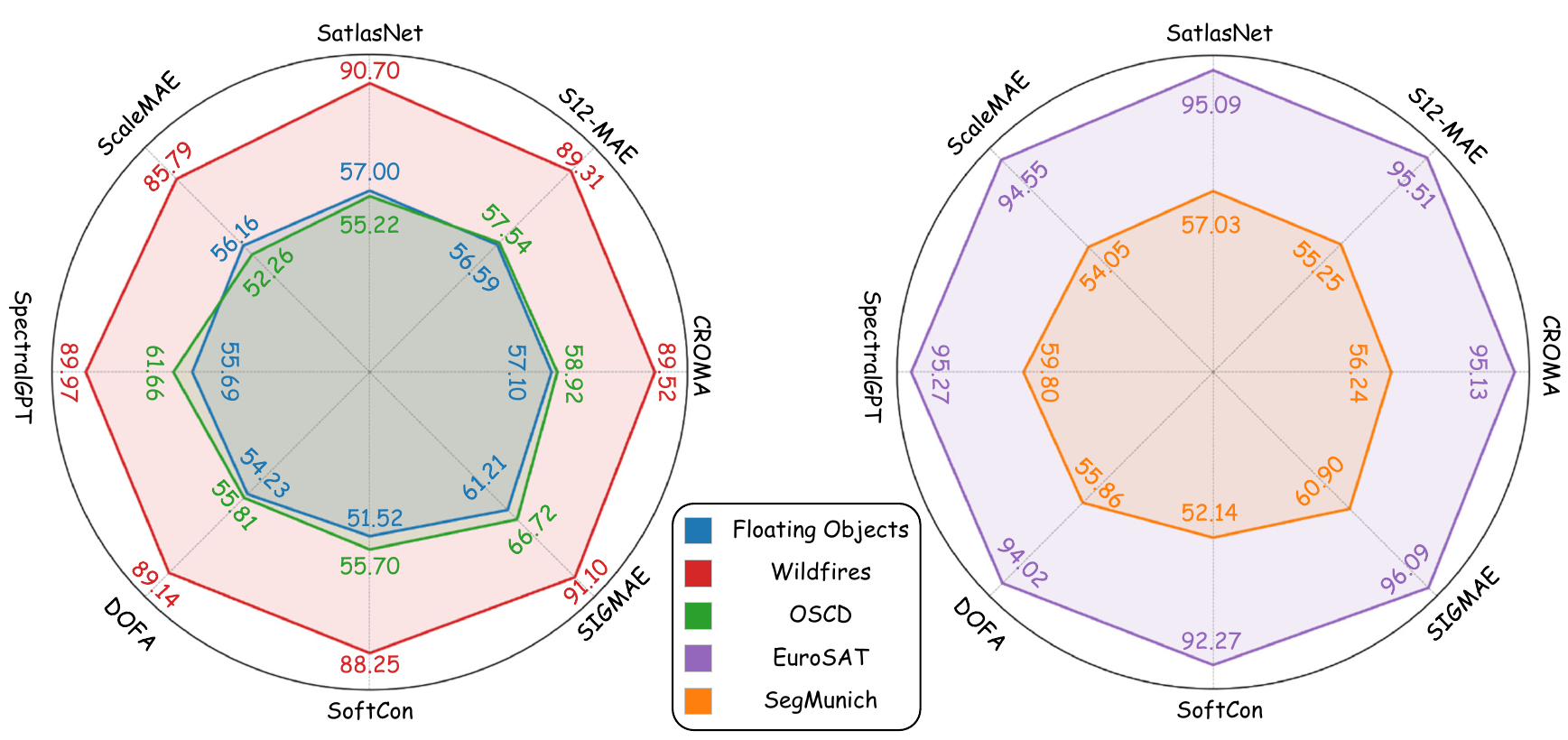}
    \caption{Performance comparison of remote sensing foundation models across five diverse datasets, where our SIGMAE achieved superior generalization capability. }
    \label{fig:spider}
\end{figure}

Recent studies have demonstrated that masked image modeling (MIM) pretraining using \textit{Masked Autoencoders} (MAE) models \citep{he2022masked} is highly effective for enhancing image representation learning. The primary principle is to reconstruct the missing regions of an input image by utilizing compressed representations learned from visible patches through an encoder–decoder framework. Such a mechanism enables the model to capture contextual dependencies across image patches, which are crucial for various vision tasks. In the remote sensing domain, recent efforts have further extended MAE by introducing spectral-aware 3D Transformers \citep{10490262}, multi-scale modeling strategies \citep{noman2024rethinking}, and spatial–temporal embeddings \citep{li2024s2mae,cong2022satmae}, in order to more effectively capture the spatial and spectral properties of multispectral imagery.
Despite its effectiveness in mining patterns from unlabeled data, MAE-based remote sensing image pretraining still faces several challenges that differ markedly from those in natural images. 
\begin{enumerate}
\item Natural images used for MAE pretraining generally have clearer object boundaries and simpler backgrounds, which facilitate the learning process. In contrast, the targets in remote sensing images often exhibit semantic dispersion due to vague contours with complex and heterogeneous backgrounds. 

\item The image reconstruction performance of MAE is highly related to the semantic components within the image \citep{li2022semmae,chen2023improving}.
Since the entire training process lacks semantic knowledge and is inherently uncontrollable, MAE tends to learn general representations, and it is difficult to explicitly construct meaningful and semantic hints for masked modeling.

\item Vision Transformer-based segmentation models struggle to learn discriminative representations with limited labeled multispectral data due to their high parameter and token counts and computational demands \citep{10490262}. Moreover, most downstream tasks focus on common categories such as buildings, vegetation, and water, while complex targets with weak and diverse spectral signatures in moderate-resolution images remain underexplored.
\end{enumerate}

To remedy these issues, we propose a novel spectral index-guided masked autoencoder (SIGMAE) for multispectral remote sensing image pretraining.
Instead of the widely used random sampling strategy for patch masking, a dynamic masking method is proposed by leveraging remote sensing spectral indices as domain knowledge to guide the training process and enhance the model’s ability to capture informative spatial-spectral properties. 
The incorporation of domain knowledge into the pretrained model enhances its feature discriminative capability and facilitates its adaptation to a wide range of image interpretation downstream tasks.
The main contributions of this work are as follows:
\begin{itemize}
\item A dynamic masking strategy that integrates remote sensing spectral indices as prior knowledge is developed to guide the masking process to focus on regions with rich spatial-spectral information, while facilitating discriminative representation learning.
\item By adopting a curriculum learning manner and dynamically balancing the impacts of informative and less-informative patches in the masking process, SIGMAE enhances the model’s ability to reason about semantic and global structural relationships during reconstruction while preserving variability to mitigate overfitting.
\item The proposed approach achieved remarkable performance on various downstream tasks while requiring relatively few parameters and limited pretraining data.
\end{itemize}

\section{Related Works}\label{sec2}
\subsection{Remote Sensing Image Representation Learning}
In remote sensing semantic segmentation tasks, Vision Transformer (ViT) and Swin Transformer (SwinT) have dominated due to their long-range context modeling capability by leveraging self-attention mechanisms, which allow them to focus on relevant areas of an image irrespective of their spatial distance \citep{10490262,10458980}. 
The ability of Transformers to effectively handle global contexts and maintain spatial relationships has been explored to capture multi-dimensional dependencies across spectral, spatial, and temporal dimensions for object extraction from multispectral images \citep{yuan2022sits, schiller2024forest}. 
To address the challenges of large-scale annotations for model training, the pretraining and fine-tuning paradigm has drawn sustained attention, with a large unlabeled set for self-supervised pretraining and then adapted to a specific task using a limited set of labeled data for fine-tuning \citep{tao2023self,mendieta2023towards,bastani2023satlaspretrain}.
Pretraining is typically conducted in a self-supervised learning manner, with methods such as contrastive learning \citep{11134807,guo2024skysense,10604292,wu2025semantic} and masked image modeling \citep{he2023ast}. 
Self-supervised learning reduces dependence on labeled datasets by pretraining models on unlabeled data through intrinsic data patterns. It enables the model to learn robust representations, which can be fine-tuned with a few labeled samples to achieve strong performance on downstream tasks \citep{9875399, xu2024self}.

\subsection{MAE-based Pretraining}

% \subsection{MAE in Remote Sensing}
% Most existing pretraining methods in remote sensing are based on the MAE framework. 
Recently, MAE has been investigated for self-supervised pretraining on large-scale, unlabeled remote sensing datasets.
For high-resolution optical images, RVST \citep{wang2022advancing} applied MAE for pretraining large ViTs and introduced rotated, variable-size window attention to handle the large and arbitrarily oriented objects in remote sensing images. Moreover, RingMo \citep{sun2022ringmo} modified the MAE masking strategy by reversing some pixels in the masked patches to retain small objects. Furthermore, multi-scale pretraining has been developed considering the scale variation nature of remote sensing data by introducing cross-scale consistency constraints \citep{reed2023scale,tang2024cross}. For multispectral image modeling, SatMAE \citep{cong2022satmae} focused on the spatial-temporal embedding of multitemporal and multispectral satellite images. For hyperspectral images, SpectralGPT \citep{10490262} introduced a novel 3D generative transformer framework, while spatial-spectral properties have been explored in \citep{ibanez2022masked} and \citep{lin2023ss}, respectively. 
SoftCon \citep{softcon} introduces a multi-label supervision framework with contrastive learning and a Siamese masking strategy to learn cross-scene soft similarities.
More recently, multimodal and multitask pretraining of Vision Transformers have been investigated for remote sensing images \citep{fuller2024croma,han2024bridging,10547536,li2025fleximo}.
Our work differs from the aforementioned approaches by enhancing the masking strategy in MAE through the integration of remote sensing spectral indices, with a specific focus on complex target segmentation tasks.

\subsection{Spectral Knowledge Guided Representation Learning}
To integrate domain knowledge into semantic segmentation models, Li et al. \citep{li2022dkdfn} incorporated spectral indices such as NDBI, NDWI, and NDVI into the training process by introducing an additional decoder designed to reconstruct these indices and guide model learning.
% an additional decoder with a knowledge reconstruction loss for domain knowledge–guided segmentation, where spectral indices such as NDBI, NDWI, and NDVI were incorporated as representative domain knowledge.
Alternatively, these indices can be treated as auxiliary input features and fused into multi-channel remote sensing composites \citep{tao2022msnet,audebert2018beyond}.
In the context of pretraining, FG-MAE \citep{10766851} employed multiple image features---such as edges, gradients, and normalized indices—as reconstruction targets, rather than the raw images.
Nevertheless, explicitly reconstructing spectral indices or embedding them into the encoder during pretraining 
inevitably introduces substantial computational overhead and increases the complexity of training.
In this article, we propose a more efficient way to incorporate the domain knowledge into the pretrained models.

\begin{figure*}
    \centering
\includegraphics[width=1\linewidth]{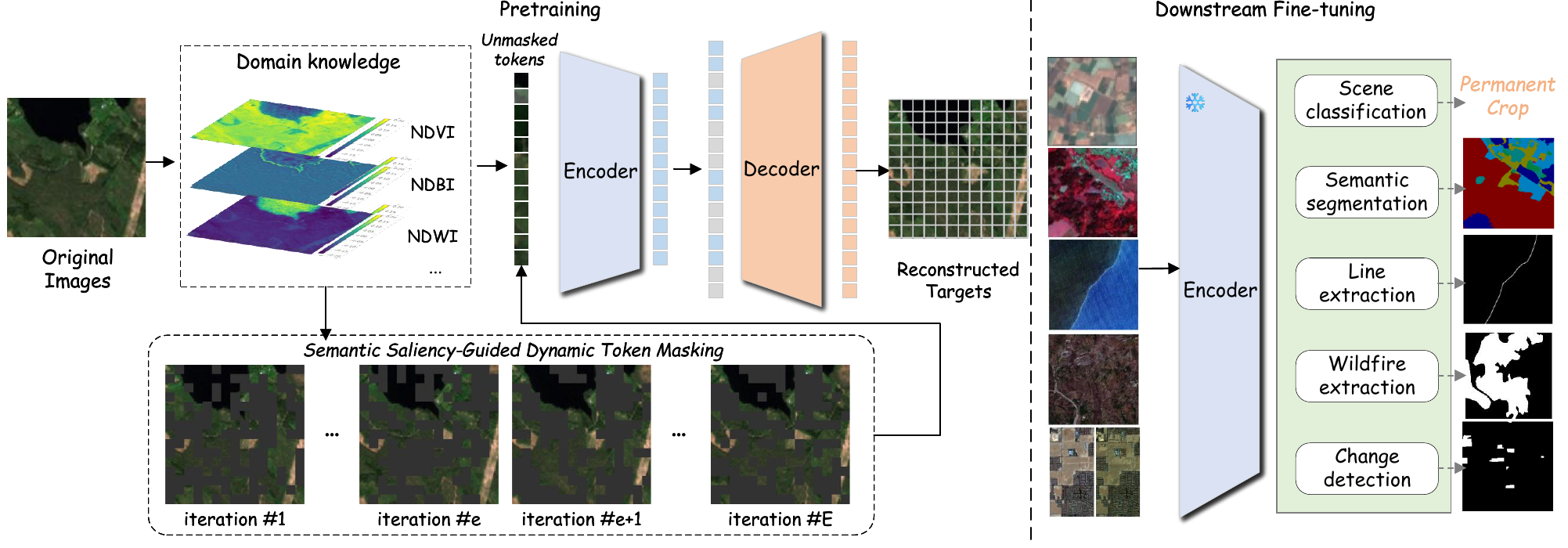}
\vspace{-8pt}
    \caption{{Overview of the proposed SIGMAE framework: an asymmetric encoder–decoder architecture equipped with Semantic Saliency-guided Dynamic Token Masking (SSDTM) that uses spectral-domain priors to adaptively select informative regions and enhance feature discriminability during reconstruction.}}
    \label{frame1}
\end{figure*}

\section{Methods}\label{sec3}
\subsection{{SIGMAE Overview}}
SIGMAE adopts an asymmetric encoder-decoder structure grounded in Vision Transformers (ViT) \citep{dosovitskiy2020image}. In this framework, the encoder is designed to process only the visible subset of the input image and learn compact feature embeddings, while the decoder reconstructs the full image content by estimating the pixel values of the masked areas.
In this study, we propose a dynamic semantic-guided masking strategy for selecting input tokens with spatially varied significance, as shown in Fig.~\ref{frame1}.

\subsubsection{Patchify}
Let the input image be represented as \( \mathcal{I} \in \mathbb{R}^{C \times H \times W} \), where \( C \) denotes the number of channels, and \( H \) and \( W \) represent the spatial dimensions. The image is first partitioned into non-overlapping patches, resulting in a sequence \( {z} \in \mathbb{R}^{L \times P^2 C} \), where each patch has size \( P \times P \), and the number of patches is given by \( L = \frac{H}{P} \cdot \frac{W}{P} \). These patches are embedded into a \( D \)-dimensional latent space through a linear projection \( f_{proj} : \mathbb{R}^{P^2 C} \rightarrow \mathbb{R}^D \), producing patch tokens of the input image \( \mathcal{Z} = \{{z_1}, {z_2}, \dots, {z_L}\} \).

% To achieve this, the algorithm uses a noise ratio that decreases over time, which allows for a smooth transition from exploration to exploitation.

\subsubsection{Domain Knowledge Embeddings}
Remote sensing spectral indices have been proven effective as prior domain knowledge to enhance the image representation capability \citep{li2022dkdfn,10766851}.
Typical spectral indices derived from remote sensing images are incorporated into the pretraining to exploit the distinctive properties of different spectral bands, providing valuable prior domain knowledge.
Following previous works, the three most commonly used indices, NDVI, NDWI, and NDBI, are adopted in this study, which are highly responsive to vegetation, water bodies, and built-up areas, respectively.

To achieve domain knowledge embeddings, the spectral index values are first calculated from the input tensor using a patch-based approach. Let \( \psi \in \mathbb{R}^{K \times H \times W} \) represent the concatenated spectral index tensor, where $K$ is the number of indices. After that, we can compute the patch-level index with a patch size of  \( P \times P \), resulting in domain knowledge embeddings  \( \mathcal{A} \in \mathbb{R}^{K \times P^2 \times L} \).

% We compute the average spectral index values over patches of size \( P = 16 \), which can be expressed by $AvgPool_{P\times P}(\psi)$, 
% resulting in the domain knowledge embeddings \( \mathcal{A} \in \mathbb{R}^{K \times L} \), where \( L \) is the number of tokens. This step flattens the spectral index values into a set of tokens, which are then used for dynamic masking.

\subsubsection{Semantic Saliency-Guided Dynamic Token Masking (SSDTM)}
Normally, for $L$ tokens in $\mathcal{Z}$, a random masking strategy is applied such that a portion \( p_m \) of the tokens is concealed. The remaining \( (1 - p_m)L \) visible tokens, enriched with positional encodings, are forwarded to the encoder to extract meaningful representations.   
In comparison, SIGMAE adopts a dynamic token masking strategy through the dynamic adjustment of semantically rich patches and informative ones during the training process.

After that, we calculate the mean $\boldsymbol{\mu}=\{\mu^{k}(\mathcal{A})\}^{K}_{k=1}$ and standard deviation $\boldsymbol{\sigma}=\{\sigma^{k}(\mathcal{A})\}^{K}_{k=1}$ of knowledge embeddings within each patch.
% Specifically, the mean reflects the semantic certainty of a patch, where higher values imply richer semantic information, making the patch more informative and showing stronger class discriminability.
{Specifically, the mean reflects both the dominant land cover and semantic certainty of a patch, where the sign determines the surface type while higher values imply richer semantic information, making the patch physically interpretable and showing stronger class discriminability.}
The standard deviation characterizes the heterogeneity within a patch, representing the reconstruction difficulty and measuring the degree of information dispersion within the patch.
% On this basis, we utilize a Semantic Saliency Measurement (SSM) to measure the semantic importance of each patch, which is defined as follows:
{On this basis, considering the spectral polarity characteristic of remote sensing indices where both positive and negative values convey high semantic importance,we utilize a Semantic Saliency Measurement (SSM) to measure the semantic importance of each patch, which is defined as follows:}
% \begin{equation}
% {Q}(\mathcal{A}_i) = \frac{1}{K}\sum_{k}\frac{\mu^k(\mathcal{A}_i)}{\sqrt{(\sigma^k(\mathcal{A}_i))^2+\epsilon}},
% \end{equation}
% \begin{equation}
% Q(\mathcal{A}_i) = 
% \frac{1}{K}\sum_{k}\frac{\left|\mu^k(\mathcal{A}_i)\right|}
% {\sqrt{(\sigma^k(\mathcal{A}_i))^2+\epsilon}},
% \end{equation}
\begin{equation}
Q(\mathcal{A}_i) = 
\frac{1}{K}\sum_{k}
\frac{\mu^k\!\left(\lvert \mathcal{A}_i\rvert\right)}
{\sqrt{(\sigma^k(\mathcal{A}_i))^2+\epsilon}},
\end{equation}
where $\epsilon$ is a small constant used to avoid division by zero.
A higher SSM indicates that the patch contains richer and more discriminative semantic information with lower internal heterogeneity, thus being more suitable for reconstruction and model training. 
{Conversely, a lower SSM implies sparse semantic information and high internal heterogeneity, resulting in greater reconstruction difficulty. Nevertheless, these patches contribute significantly to providing discriminative features for model training.}

% \textcolor{red}{This comprehensive statistical characterization, as visualized in Fig. \ref{analysis}, forms the foundation of our masking strategy. By leveraging the distribution of remote sensing indices and semantic richness, the derived SSM acts as a guiding metric to modulate the content-aware masking schedule. This ensures the model adaptively prioritizes patches with high semantic saliency, thereby enhancing feature representation learning.}

% \begin{equation}
% S =
% \begin{cases}  
%   2 \rho  \frac{1}{K}\sum_{k}\frac{\mu^k(\mathcal{A})}{\sigma^k(\mathcal{A}_i)+\epsilon} + (1 - \rho) \, \epsilon, & v \ge 0 \\[1mm]
%   (1-2 \rho)  \frac{1}{K}\sum_{k}\frac{\mu^k(\mathcal{A})}{\sigma^k(\mathcal{A}_i)+\epsilon} + (1 - \rho) \, \epsilon, & v < 0
% \end{cases}
% \end{equation}

% \begin{equation}
% S =
% \begin{cases}  
%   2 \rho \mathcal{Q}(\mathcal{A}_i)  + (1 - \rho) \epsilon, & v \ge 0 \\[1mm]
%   (1-2 \rho)  \mathcal{Q}(\mathcal{A}_i) + (1 - \rho) \epsilon, & v < 0
% \end{cases}
% \end{equation}
At each epoch, the dynamic masking score $S$ for each token is computed by combining the SSM values and random noise, which can be expressed by:
% \begin{equation}
% S(\mathcal{A}_i; e) = (1 - 2 \gamma(e)) Q(\mathcal{A}_i) + (1 - \gamma(e)) \nu_i,\label{score}  
% \end{equation}
% \begin{equation}
% S(\mathcal{A}_i,e) =
% \begin{cases}  
%   (1-2 \gamma(e) \, {Q}(\mathcal{A}_i)  +2\gamma(e))\,\nu, & 0 < \gamma(e) \le 0.5 \\[1mm]
%   -\gamma(e)\, {Q}(\mathcal{A}_i) + (1 - \gamma(e))\,\nu, & 0.5 < \gamma(e) \le 1
% \end{cases}\label{score}
% \end{equation}
\begin{equation}
{\footnotesize
S(\mathcal{A}_i,e) =
\begin{cases}  
  (1-2 \gamma(e))\, Q(\mathcal{A}_i) + 2\gamma(e))\,\nu, 
  & 0 < \gamma(e) \le 0.5 \\[1mm]
  -\gamma(e)\, Q(\mathcal{A}_i) + (1 - \gamma(e))\,\nu, 
  & 0.5 < \gamma(e) \le 1
\end{cases}
}
\label{score}
\end{equation}
where \( \gamma(e)=e/E\) is a dynamic scaling factor that evolves over epochs indexed by $e$, where $E$ is the total number of epochs and $\nu \sim \mathcal{U}(0, 1)$ represents the random noise, sampled uniformly from the range \( [0, 1] \). 
% The whole dynamic token masking adopts a curriculum learning manner. It explicitly defines sample difficulty based on the mean and standard deviation of knowledge embeddings, and gradually guides the model from easier samples that are semantically homogeneous and easier to reconstruct, to harder samples containing richer details and harder to reconstruct.

{By adopting a curriculum learning paradigm, our method explicitly defines sample difficulty based on the mean and standard deviation of knowledge embeddings. By dynamically modulating the focus through $\gamma(e)$, this strategy establishes a distinct `Simple-to-Random-to-Hard' progression: guiding the model from capturing dominant semantic features, through a stochastic transition, to focusing on subtle details. This content-aware schedule prevents early overfitting while ensuring later robustness, thereby demonstrating superior feature discriminability compared to static stochastic strategies.}

% \textcolor{red}{By dynamically modulating the masking focus through the scaling factor $\gamma(e)$, our method establishes a distinct 'Simple-to-Random-to-Hard' progression. Specifically, the model first prioritizes capturing dominant semantic features, transitions through a stochastic phase, and finally shifts attention to subtle details. This content-aware curriculum prevents overfitting to simple patterns in the early stages while ensuring robustness against complex textures in the later stages, thereby demonstrating superior feature discriminability compared to static stochastic strategies.}
% At each epoch, the dynamic masking
% score $S$ for each token is computed by combining the spectral index values and random noise, which can be expressed by:
% \begin{equation}
%     S(\mathcal{A}_i)=(2\rho-1)\frac{1}{K}\sum_{k}\frac{\mu^k(\mathcal{A}_i)}{\sigma^k(\mathcal{A}_i)+\epsilon},
% \end{equation}
% where $\epsilon$ is a set to avoid the NaN values.

To determine which tokens should be masked, we rank the tokens based on their dynamic scores, as follows:
\begin{equation}
\mathbf{T}^{S}=\left\{ i \;\middle|\; i \in \mathrm{top}_{\left\lfloor p_m \times L \right\rfloor}  \left( \left\{ S(\mathcal{A}_i) \right\}_{i=1}^{L} \right)  \right\},
\end{equation}
where $\mathbf{T}^{S}$ means the selected tokens to be masked and the binary mask can be obtained by:
\begin{equation}
\mathcal{M}_{\text{binary}}= \mathbb{I}\Big\{ (h,w) \in \bigcup_{i \in \mathbf{T}^S} {z}_i \Big\}, 
\end{equation}
where $\mathcal{M}_{\text{binary}}$ is a binary mask and
$\mathbb{I}$ is an indicator function.

\subsubsection{Pretraining}
To recover the original image, the decoder takes all \( L \) token positions as input: the visible ones are inserted back into their original locations, while the masked positions are filled with a trainable mask token. Positional encodings are again incorporated to guide the reconstruction process. The decoder then produces the output \( \hat{\mathcal{I}} \in \mathbb{R}^{C \times H \times W} \), aiming to reconstruct the original image as accurately as possible.

The model is trained by minimizing the reconstruction loss between the original image \( \mathcal{I} \) and the predicted reconstruction \( \hat{\mathcal{I}} \), focusing only on the masked pixels. The reconstruction loss \( \mathcal{L}_{\text{MAE}} \) is typically computed using the mean squared error (MSE) between the original and reconstructed values at the masked positions:
\begin{equation}
\mathcal{L}_{\text{MAE}} = \frac{1}{|\mathcal{M}|} \sum_{h,w} \mathcal{M}_{h,w} \cdot \left( \mathcal{I}_{h,w} - \hat{\mathcal{I}}_{h,w} \right)^2,
\end{equation}
where \( \mathcal{M}_{h,w} \) is the mask value at position \( (h, w) \), and \( |\mathcal{M}| \) is the total number of masked pixels (i.e., where \( \mathcal{M}_{h,w} = 1 \)). This ensures that only the masked pixels contribute to the loss computation.
The details of the pretraining process are illustrated in Algorithm~\ref{alg:sigmae}.

\begin{algorithm}[htb]
\caption{SIGMAE Pretraining}\label{alg:sigmae}
\KwIn{$\mathcal{I}$: input image; $\psi$: spectral index tensor; $p_m$: mask ratio; $P$: patch size; $E$: total number of epochs; $\eta$: learning rate; $\theta$: network parameters}
\KwOut{Pretrained network parameters $\theta$}

\For{$e = 1, \dots, E$}{
    Split $\mathcal{I}$ into non-overlapping patches: $\mathcal{Z} \gets \text{Patchify}(\mathcal{I}, P)$\;
    Spectral knowledge embedding: $[\mu;\sigma] \gets (\psi, P)$\;
    Calculate masking score ${S}$ according to Eq.~\ref{score}\;
    Generate binary mask: 
    $\mathbf{T}^{S} \leftarrow \left\{ \mathcal{A}_i \;\middle|\; i \in \mathrm{top}_{\left\lfloor p_m \times L \right\rfloor}  \left( \left\{ S(\mathcal{A}_i) \right\}_{i=1}^{L} \right)  \right\}$\;
    Obtain visible tokens: $\mathcal{Z}_{v} \gets \mathcal{Z} \setminus \mathbf{T}^{S}$\;
    Encode visible tokens: $\mathbf{h}_e \gets \text{Encoder}(\mathcal{Z}_{v})$\;
    Decode full sequence: $\mathbf{h}_d \gets \text{Decoder}(\mathbf{h}_e, \mathcal{M}_\mathrm{binary})$\;
    Predict original patches: $\hat{\mathcal{Z}} \gets \text{ProjectionHead}(\mathbf{h}_d)$\;
    Update parameters: $\theta \leftarrow \theta - \eta \cdot \nabla_{\theta} \mathcal{L}_\mathrm{MAE}$\;
}
\end{algorithm}

\begin{figure*}
    \centering
\includegraphics[width=1\linewidth]{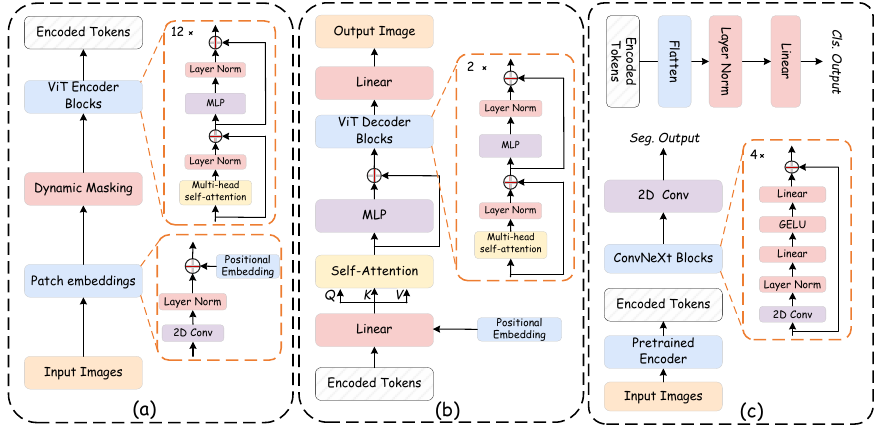}
    \caption{Details of network structures of the (a) encoder, (b) reconstruction decoder and (c) for downstream tasks.}
    \label{struc}
\end{figure*}

\subsection{{Network Structures}}
The details of network structures in SIGMAE are illustrated in Fig.~\ref{struc}. 
A ViT-based autoencoder is adopted with an encoder and a reconstruction decoder. 

First, $16 \times 16$ image patches are projected into tokens with the appropriate Transformer dimension. These projected patches are then concatenated into a token sequence.
Pretrained weights can be directly loaded into a standard ViT by adjusting the input projection accordingly. After the linear projection, 2D sine--cosine positional embeddings are added before dynamic masking.

The decoder receives the full set of visible tokens as input.
These visible tokens are jointly decoded along with a set of mask tokens, which act as placeholders allowing the decoder to reconstruct the masked patches from the visible ones. As illustrated in Fig.~\ref{struc}, the decoding process begins with a linear projection layer that aligns the dimensionality of the encoder output tokens with that of the decoder. Sine–-cosine positional embeddings are then added to preserve spatial structure. The enriched tokens are subsequently fed into a stack of self-attention and MLP layers, followed by multiple Transformer blocks, which progressively refine the feature representations.

% Positional embeddings (sine-cosine) and learned modality embeddings are then added to the decoder inputs. 

% and fed into the Transformer encoder

% First, 16$\times$16 image patches are projected to tokens with the correct Transformer dimension. Projected patches are then concatenated into a sequence of tokens and given as input to the Transformer encoder. The Transformer encoder is adopted from ViT \citep{dosovitskiy2020image}. pretrained weights can be used directly in a standard ViT by loading the desired input projection.
% We add 2D sine-cosine positional embeddings after the linear projection.
% These visible tokens are decoded jointly with a set of mask tokens, which serve as placeholders for the decoders to write the reconstructed patches and to reconstruct the masked-out tokens from the visible tokens. As shown in Fig.~\ref{frame2}, each decoder has a linear projection layer to adapt the outputs from the encoder to the decoder dimension. After this linear projection, we add both sine-cosine positional embeddings and learned modality embeddings to the decoder inputs. This is then followed by a cross-attention layer, an MLP, and two Transformer blocks.

\subsection{Dataset for Pretraining}
The BigEarthNet-S2 \citep{8900532} dataset serves as a large-scale benchmark for foundation model pretraining. It contains $590,326$ image patches of size $120 \times 120$ extracted from Sentinel-2 satellite imagery collected during summer over Austria, Belgium, Finland, Ireland, Lithuania, Serbia, and Switzerland. Each patch is annotated with multiple land-cover labels derived from the CORINE Land Cover Map. 
For pretraining, ten Sentinel-2 bands were employed, excluding the two 60~m resolution bands that are primarily intended for atmospheric correction, cirrus detection, and cloud screening. The remaining 20~m resolution bands were upsampled to 10~m using nearest-neighbor interpolation to ensure spatial consistency across all bands.
% \textcolor{red}{As shown in Fig.~\ref{analysis}, the distribution of remote sensing indices, semantic richness, and the resulting Semantic Saliency Measurement (SSM) are key elements used for modulating the content-aware masking schedule during pretraining.}

\subsection{Fine-Tuning}
To support multiple downstream tasks, the unified token representation is routed to task-specific heads. For image-level classification, the token sequence is flattened and normalized before being mapped to category logits through a linear classifier, enabling efficient global prediction. To fine-tune the pretrained encoder for the image segmentation task, ConvNeXt \citep{liu2022convnet}, constructed from a pure convolutional net, is adopted as the segmentation head, as shown in Fig.~\ref{struc}. Leveraging patch-wise stems, inverted bottlenecks, and large kernels, ConvNeXts achieves competitive accuracy and scalability compared to transformers. More specifically, the output tokens are first passed through a linear projection to increase their dimensionality. After processing through ConvNeXt blocks, the feature map is upsampled back to the original image resolution using bilinear interpolation.
For change detection, the two input images are fed through the same pretrained encoder, and their multi-level features are extracted and fused. Specifically, the corresponding intermediate representations from the four encoder stages are concatenated and subsequently provided to the decoder as its input.

\begin{table}[htp]
\caption{Summary of datasets used in the downstream tasks}
\centering
\begin{tabular}{lcccc}
\toprule
Dataset & Task & Image Size & Fine-tuning & Testing \\
\midrule
FOD & Seg.        & $224 \times 224$ & 6,087  & 3,704 \\
Wildfire Detection        & Seg.        & $224 \times 224$ & 2,263  & 4,053 \\
EuroSAT                   & Cls.   & $64 \times 64$   & 21,600 & 5,400 \\
SegMunich                 & Seg.     & $128 \times 128$ & 39,402 & 9,846 \\
OSCD                      & CD  & $224 \times 224$ & 7,160    & 3600 \\
\bottomrule
\end{tabular}
\label{tab:datasets}
\end{table}

\section{Experiments}\label{sec4}
\subsection{Benchmark Datasets}
% Sentinel-2 is an Earth observation mission under the Copernicus Programme. It consists of two twin satellites, Sentinel-2A and Sentinel-2B, which capture multispectral imagery across 12 spectral bands with spatial resolutions ranging from 10 m to 60 m. The mission offers a revisit time of 10 days at the equator with one satellite, which improves to 5 days when both satellites are operational, and further reduces to 2–3 days at mid-latitudes.

To assess the effectiveness of the pretrained foundation models, we conduct experiments on five datasets covering four downstream remote sensing tasks: scene classification, object extraction, semantic segmentation, and change detection. Beyond commonly used datasets, we also consider more challenging targets, such as floating objects and wildfire-affected areas, which are difficult to detect due to their weak and diverse spectral signatures in moderate-resolution imagery. The detailed settings of datasets in the downstream tasks are listed in Table~\ref{tab:datasets}.

% \begin{figure*}
%     \centering
% \includegraphics[width=1\linewidth]{figs/data_analysis.pdf}
%     \caption{\textcolor{red}{
%     Illustration of a comprehensive statistical characterization of the dataset's spectral attributes and the derived patch-level information metrics. Specifically, (a), (b), and (c) sequentially illustrate the distribution of raw remote sensing indices, the semantic richness of knowledge embeddings, and the resulting Semantic Saliency Measurement (SSM) used to modulate our content-aware masking schedule.}}
%     \label{analysis}
% \end{figure*}

\subsubsection{Floating Objects Detection (FOD) Dataset}
This dataset is designed for detecting floating objects on the sea surface using globally available medium-resolution Sentinel-2 imagery \citep{mifdal2021towards}. It includes six categories of floating objects: plastic, pumice, seafoam, seawater, seaweed, and timber.
This large-scale global data has a high diversity of floating objects, containing a dataset covering several coastal regions, such as Panama, Lagos, and Shengsi. In the experiments, all images were clipped into patches with a size of $224 \times 224$ pixels with a 64-pixel overlap. The training and test sets were split according to the coastal regions with a ratio of 2:1.

\subsubsection{Wildfire Detection Dataset}
The wildfire detection dataset \citep{arnaudo2023robust} includes Sentinel-2 imagery associated with wildfires, along with delineation masks, covering June 2017 to April 2023.
The dataset includes $73$ forest wildfires recorded between 2017 and 2019 across different regions in Europe. It covers a total area of approximately $19,000$ $\text{km}^2$, featuring diverse terrain and morphological characteristics.
Therefore, this dataset provides a robust benchmark for wildfire-related segmentation tasks. In the experiments, all images were clipped into patches of size $224 \times 224$ without overlaps. In the experiments, a limited number of samples are employed for fine-tuning, whereas a larger set is reserved for testing.

\subsubsection{EuroSAT \citep{helber2019eurosat}}
The EuroSAT dataset contains 27,000 Sentinel-2 satellite images collected from 34 European countries, each spanning $64 \times 64$ pixels with 13 spectral bands. The images are labeled into 10 land use classes, including Industrial Buildings, Residential Buildings, Annual Crop, Permanent Crop, Sea \& Lake, Herbaceous Vegetation, Highway, Pasture and Forest, with each class comprising between 2,000 and 3,000 samples. The dataset is partitioned into training and test subsets using an 8/2 split.

\subsubsection{SegMunich \citep{10490262}}
The dataset is constructed as a 10-band best-pixel composite with a spatial resolution of 10 m, covering an area of 3,847 × 2,958 pixels over a three-year period up to April 2020. It provides a segmentation mask that delineates 13 land-cover classes within the Munich metropolitan region, including categories such as arable land, pastures, forests, surface water, shrubland, and wetlands. The imagery is partitioned into patches of $128 \times 128$ pixels with 50\% overlap, and subsequently divided into training and validation sets in an 8:2 ratio.

\subsubsection{OSCD \citep{daudt2018urban}}
The Onera Satellite Change Detection (OSCD) dataset is a benchmark designed for evaluating urban change detection methods using multispectral satellite imagery. It contains 24 pairs of co-registered Sentinel-2 images, each spanning 13 spectral bands, acquired between 2015 and 2018 from diverse regions worldwide, including Europe, the Middle East, Asia, the USA, and Brazil.
Each image pair has a spatial size of 600 × 600 pixels, with resolution varying from 10 m to 60 m depending on the spectral band. Pixel-level change annotations are provided for 14 training pairs and 10 test pairs, highlighting urban transformations such as the development of new buildings or roads. All images were clipped into patches of size $224 \times 224$ with 40\% overlaps.

\begin{table*}[htp]
\caption{Foundation models used in benchmarking, with architecture, pretraining data, strategy, size, patch volume, and release year.}
\label{table:FMs}
\centering
\footnotesize % 建议使用 footnotesize，比 scriptsize 稍微大点更易读
% \setlength{\tabcolsep}{3pt} % 适当间距
% \renewcommand{\arraystretch}{1.2}
% 使用普通的 tabularx 逻辑，不搞复杂的 hsize 权重
\begin{tabularx}{\textwidth}{@{} l c X >{\centering\arraybackslash}p{1.8cm} c c @{}}
\toprule
\textbf{Model} & \textbf{Arch.} & \textbf{Pretrained EO Data} & \textbf{\makecell{Learning\\Strategy}} & \textbf{\makecell{Params\\(M)}} & \textbf{\makecell{Patch\\Volume}} \\
\midrule
CROMA \textcolor{gray}{\scriptsize(NIPS'24)} & ViT & SSL4EO-S12 {\scriptsize \citep{wang2023ssl4eo}} & Contrastive & 396.13 & 3M \\
SatlasNet \textcolor{gray}{\scriptsize(ICCV'23)} & Swin-T & SatlasPretrain {\scriptsize \citep{bastani2023satlaspretrain}} & Supervised & 128.57 & 856K \\
S12-MAE \textcolor{gray}{\scriptsize(GRSM'23)} & ViT & SSL4EO-S12 {\scriptsize \citep{wang2023ssl4eo}} & MIM & 61.99 & 3M \\
ScaleMAE \textcolor{gray}{\scriptsize(ICCV'23)} & ViT & fMoW-RGB {\scriptsize \citep{christie2018functional}} & MIM & 396.21 & 363.6K \\
SpectralGPT \textcolor{gray}{\scriptsize(TPAMI'24)} & ViT & fMoW-S2, BigEarthNet {\scriptsize \citep{8900532}} & MIM & 614.75 & 1.47M \\
SoftCon \textcolor{gray}{\scriptsize(TGRS'24)} & ViT  & SSL4EO-S12 {\scriptsize \citep{wang2023ssl4eo}} & Contrastive & 242.19  & 3M \\
DOFA \textcolor{gray}{\scriptsize(arXiv'24)} & ViT & DOFA {\scriptsize \citep{xiong2024neural}} & MIM & 178.20 & 8.08M \\
\midrule
\textbf{SIGMAE (Ours)} & ViT & BigEarthNet {\scriptsize \citep{8900532}} & MIM & \textbf{118.90} & \textbf{54.9K} \\
\bottomrule
\end{tabularx}
\end{table*}

\subsection{Experimental Setup}
The proposed approach was compared with several state-of-the-art foundation models, including SatlasNet \citep{bastani2023satlaspretrain}, S12-MAE from SSL4EO-S12 \citep{wang2023ssl4eo}, CROMA \citep{fuller2024croma},  SpectralGPT \citep{10490262}, DOFA \citep{xiong2024neural}
and ScaleMAE \citep{reed2023scale} to evaluate its performance. 
The architecture, pretraining data, strategy, model size, and patch volume of the compared foundation models are summarized in Table~\ref{table:FMs}. Most of these models adopt transformer-based architectures, with ViT \citep{dosovitskiy2020image} being a representative choice. Their sizes range from 61.99M to 614.75M parameters, and the patch volumes vary from 54.9K to 8.08M, reflecting different levels of spatial granularity considered during pretraining. Specifically, S12-MAE and CROMA are pretrained on 3 million samples from Sentinel-2 and Sentinel-1 imagery, while SatlasNet is trained in a fully supervised manner with 856K samples. SpectralGPT, on the other hand, leverages fMoW-S2 \citep{christie2018functional} and BigEarthNet \citep{8900532} with the largest model size among them. In contrast, SIGMAE is characterized by a comparatively compact model size and the minimal patch volume, underscoring its efficiency in representation learning. For all baseline methods, we employ the publicly released weights before fine-tuning on the labeled datasets.

\begin{table}[!htbp]
    \centering
    \caption{Parameter Settings in the Pretraining and Fine-tuning.}
    \begin{tabular}{lll}
        \toprule
        \textbf{Hyperparameters} & \textbf{Pretraining} & \textbf{Fine-tuning} \\
        \midrule
        Optimizer & AdamW & AdamW \\
        Base learning rate & 1e-4 & 1e-4 \\
        Weight decay & 0.05 & 0.001 \\
        Adam $\beta$ & (0.9, 0.95) & (0.9, 0.999) \\
        Layer-wise lr decay & -- & 0.75 \\
        Batch size & 900 & 20 \\
        Learning rate sched. & Cosine decay & Cosine decay \\
        Training epochs & 1000 & 50 \\
        Warmup learning rate & 1e-6 & 1e-6 \\
        Warmup epochs & 20 & 1 \\
        Non-masked tokens & 112 & -- \\
        % Task weighting & None (equal weights) & -- \\
        Input resolution & $120 \times 120$ & e.g., $224 \times 224$ \\
        % Augmentation & RandomResizedCrop & -- \\
        \bottomrule
    \end{tabular}
    \label{tab:train_settings}
\end{table}

For quantitative evaluation, five commonly used metrics were employed: overall accuracy (OA), precision, recall, F1-score, and mean intersection over union (mIoU).  
The parameter settings of the SIGMAE pretraining and fine-tuning are displayed in Table~\ref{tab:train_settings}, respectively. All models were implemented using the PyTorch framework. The experiments were conducted on an Ubuntu operating system equipped with four NVIDIA A100 Tensor Core GPUs.

\subsection{{Quantitative Analysis}}
The quantitative results presented in Tables~\ref{combined_results_all} and~\ref{segmunich_results} highlight the superior performance of the proposed SIGMAE among existing models.

% Moreover, it demonstrates promising generalization capability across diverse datasets, as shown in Fig.~\ref{fig:loss}.

\begin{table*}[!htp]
\centering
\caption{Quantitative Results on FOD, Wildfires, OSCD, and EuroSAT Datasets for Segmentation, Change Detection, and Classification Tasks (\%)}
\setlength{\tabcolsep}{1pt} 
\renewcommand{\arraystretch}{1.2}
\footnotesize 
\begin{tabular}{l cccc cccc cccc ccc}
\toprule
Model 
& \multicolumn{4}{c}{FOD} 
& \multicolumn{4}{c}{Wildfires Detection} 
& \multicolumn{4}{c}{OSCD} 
& \multicolumn{3}{c}{EuroSAT} \\
\cmidrule(lr){2-5} 
\cmidrule(lr){6-9} 
\cmidrule(lr){10-13} 
\cmidrule(lr){14-16}
& mIoU & F1 & Recall & Precision 
& mIoU & F1 & Recall & Precision 
& mIoU & F1 & Recall & Precision 
& mIoU & OA & mF1 \\
\midrule
CROMA        
& 57.10 & 62.87 & 72.68 & 59.05 
& 89.52 & 89.52 & \textbf{95.59} & 84.18 
& 58.92 & 66.98 & 76.53 & 62.80 
& 95.13 & 97.61 & 97.49 \\

S12-MAE      
& 56.59 & 62.09 & 60.58 & 64.15 
& 89.31 & 89.28 & 84.59 & 94.53 
& 57.54 & 65.09 & 76.72 & 60.90 
& 95.51 & 97.85 & 97.73 \\

SatlasNet    
& 57.00 & 62.79 & 62.62 & 62.98 
& 90.70 & 90.78 & 94.96 & 86.96 
& 55.22 & 61.58 & \textbf{87.67} & 57.39 
& 95.09 & 97.59 & 97.44 \\

ScaleMAE     
& 56.16 & 61.49 & 67.24 & 58.66 
& 85.79 & 85.25 & 82.15 & 88.60 
& 52.26 & 57.09 & 72.60 & 54.70 
& 94.55 & 97.25 & 97.14 \\

SpectralGPT  
& 55.69 & 60.68 & 74.28 & 56.93 
& 89.97 & 90.00 & 85.44 & \textbf{95.07} 
& 61.66 & 70.50 & 79.49 & 63.30       
& 95.27 & 97.94 & 97.72 \\

DOFA         
& 54.23 & 58.35 & 66.76 & 55.63 
& 89.14 & 89.09 & 84.44 & 94.29 
& 55.81 & 63.00 & 67.40 & 60.56 
& 94.02 & 96.98 & 96.86 \\

SoftCon      
& 51.52 & 53.56 & 64.61 & 52.12 
& 88.25 & 88.07 & 85.12 & 91.22 
& 55.70 & 62.40 & 78.96 & 58.36 
& 92.27 & 96.09 & 95.93 \\

\midrule
SIGMAE       
& \textbf{61.21} & \textbf{68.87} & \textbf{76.54} & \textbf{64.72} 
& \textbf{91.10} & \textbf{91.02} & 92.92 & 90.21 
& \textbf{66.72} & \textbf{76.33} & 78.29 & \textbf{74.65} 
& \textbf{96.09} & \textbf{98.09} & \textbf{97.99} \\
\bottomrule
\end{tabular}
\label{combined_results_all}
\end{table*}

Table \ref{combined_results_all} reports the quantitative results on the Floating Objects detection, Wildfire detection, OSCD and EuroSAT datasets. The results indicate that the competing methods achieve leading scores only on a few isolated metrics or datasets. For example, CROMA attains the highest Recall on Wildfires, SatlasNet achieves the best Recall on OSCD, and SpectralGPT provides the highest Precision on Wildfires. In contrast, the proposed SIGMAE consistently delivers superior performance across most metrics, achieving the highest mIoU, F1, and Precision on Floating Objects, the highest mIoU and F1 on Wildfires, and the best overall results on OSCD. These results demonstrate the generalization ability of SIGMAE across different remote sensing tasks.
For the image classification task on the EuroSAT dataset, SIGMAE obtains the best mIoU with comparable performance on the metrics, demonstrating its overall competitiveness.

\begin{table*}[!htp]
\centering
\caption{Quantitative Results on SegMunich Dataset (\%)}
\label{segmunich_results}

\setlength{\tabcolsep}{1pt} 
\renewcommand{\arraystretch}{1.2}

\footnotesize 
\newcolumntype{C}{>{\centering\arraybackslash}X}

\begin{tabularx}{\textwidth}{@{} l *{13}{C} c @{}}
\toprule
Method & \rotatebox{80}{Background} 
      & \rotatebox{80}{Arable land} 
      & \rotatebox{80}{Perm. Crops} 
      & \rotatebox{80}{Pastures} 
      & \rotatebox{80}{Forests} 
      & \rotatebox{80}{Surface water} 
      & \rotatebox{80}{Shrub} 
      & \rotatebox{80}{Open spaces} 
      & \rotatebox{80}{Wetlands} 
      & \rotatebox{80}{Mine, dump} 
      & \rotatebox{80}{Artificial veg.} 
      & \rotatebox{80}{Urban fabric} 
      & \rotatebox{80}{Buildings} 
      & mF1 \\
\midrule
CROMA        & 90.64 & 82.77 & 21.42 & 64.07 & 89.12 & 70.51 & 20.62 & 33.32 & 44.72 & 40.57 & 27.79 & 78.33 & 67.25 & 56.24 \\
S12-MAE      & 89.48 & 81.89 & 22.25 & 62.24 & 88.51 & 71.25 & 16.73 & 34.80 & 44.08 & 40.94 & 25.51 & 76.32 & 64.25 & 55.25 \\
SatlasNet    & \textbf{90.81} & 81.69 & 17.98 & 61.32 & 88.43 & 81.30 & 16.45 & 43.81 & 50.44 & 47.08 & 22.44 & 75.93 & 63.69 & 57.03 \\
ScaleMAE     & 85.12 & 81.14 & 20.04 & 60.82 & 87.62 & 70.49 & 15.77 & 29.45 & 44.29 & 39.47 & 25.52  & 77.32 & 65.57 & 54.05 \\
SpectralGPT  & 87.80 & 83.34 & 31.24 & \textbf{65.53} & 89.11 & \textbf{85.84} & 21.31 & 46.05 & 46.41 & \textbf{52.59} & 23.98 & 78.97 & 65.18 & 59.80 \\
DOFA         & 89.91 & 81.97 & 21.51 & 62.41 & 88.65 & 71.09 & 18.71 & 33.04 & 43.11 & 39.78 & 27.99 & 77.57 & 66.21 & 55.86 \\
SoftCon      & 89.77 & \textbf{88.51} &  11.21 & 64.58 & 88.89 & 62.52 & 11.76 & 24.11 & 34.70  & 30.55 & 20.12 & \textbf{82.11} & \textbf{68.93} & 52.14 \\
\midrule
\textbf{SIGMAE} & 90.14 & 81.09 & \textbf{32.71} & 63.65 & \textbf{89.37} & 85.09 & \textbf{22.26} & \textbf{48.88} & \textbf{55.15} & 50.51 & \textbf{31.31} & 77.05 & 64.47 & \textbf{60.90} \\
\bottomrule
\end{tabularx}
\end{table*}

Table \ref{segmunich_results} presents the quantitative performance of different methods on the SegMunich dataset across 13 land-cover categories, reported in terms of F1-scores. As observed, the results show that existing approaches achieve the best performance only in a few individual categories, but fail to deliver competitive results across all classes. 
For example, SatlasNet achieves the highest performance in Background and Surface water, while SpectralGPT leads in Pastures, Surface water, and Mine, dump. 
In contrast, the proposed SIGMAE consistently attains strong performance in multiple categories, including Permanent crops, Forests, Shrub, Open spaces, Wetlands, Artificial vegetation, and achieves the highest mean F1-score, demonstrating its superior overall effectiveness and robustness.

% \begin{table*}[!htbp]
% \centering
% \caption{Ablation Studies (\%)}
% % \renewcommand{\arraystretch}{2.1}
% % \setlength{\tabcolsep}{4pt}
% \begin{tabular}{lcccccccccc}
% \toprule
% Model Variants & \multicolumn{2}{c}{FOD} & \multicolumn{2}{c}{Wildfire Detection} & \multicolumn{2}{c}{EuroSAT} & \multicolumn{2}{c}{SegMunich} & \multicolumn{2}{c}{OSCD} \\
% % \cmidrule(lr){2-3} \cmidrule(lr){4-5} \cmidrule(lr){6-7} \cmidrule(lr){8-9} \cmidrule(lr){10-11}
%  & mIoU & F1 & mIoU & F1 & mIoU & mF1 & mIoU & mF1 & mIoU & F1 \\
% \midrule
% \midrule
% SIGMAE w/o SSDTM & 59.19 & 65.96 & 89.92 & 89.96 & 94.54 & 97.15 & 46.16 & 59.98 & 65.04 & 74.51 \\
% MultiMAE & 59.70 & 67.03 & 89.06 & 89.04 & 95.16 & 97.50 & 46.31 & 59.96 & 64.96 & 74.44 \\
% SIGMAE w/o $\gamma(e)$ & \textbf{61.23} & 68.78 & 90.67 & 90.73 & 94.58 & 97.18 & 46.44 & 60.16 & \textbf{66.07} & \textbf{75.70} \\
% SIGMAE (Full)  & 61.21 & \textbf{68.87} & \textbf{91.12}& \textbf{91.20} & \textbf{95.53} & \textbf{97.70} & \textbf{47.45} & \textbf{60.90} & 65.45 & 75.03 \\
% \bottomrule
% \end{tabular}
% \label{tab:ablation}
% \end{table*}

\subsection{Interpretability analysis}
\begin{figure*}
    \centering
\includegraphics[width=1\linewidth]{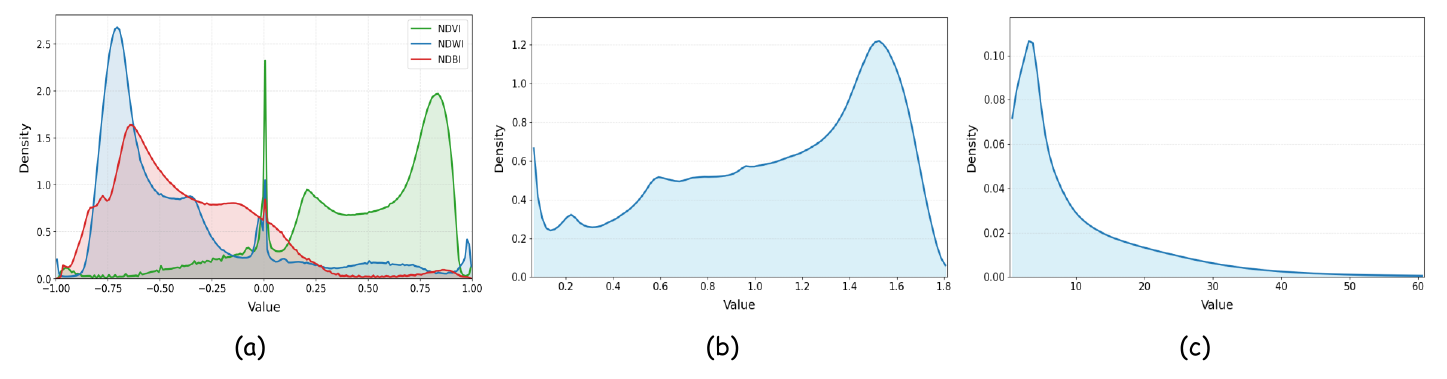}
    \caption{{Statistical characterization of spectral attributes and patch-level saliency indicators in BigEarthNet-S2 dataset. (a) Probability density functions of raw spectral indices (NDVI, NDWI, and NDBI) within the BigEarthNet-S2 dataset. (b) Distribution of the sum of absolute values derived from the indices, representing semantic richness. (c) Probability density of the resulting SSM.}}
    \label{analysis}
\end{figure*}

{This comprehensive statistical characterization, as visualized in Fig.~\ref{analysis} , forms the foundation of our masking strategy. Specifically, Fig.~\ref{analysis} (a) illustrates the distribution of raw remote sensing indices and their coupling relationships across diverse environments. To quantify the information density, the semantic richness of the knowledge embeddings is explicitly derived by summing the absolute values of these remote sensing indices as presented in Fig.~\ref{analysis} (b).By calculating the ratio of this semantic richness to the standard deviation, we derive the SSM for each patch. As illustrated in Fig.~\ref{analysis} (c), compared to the distribution of remote sensing indices in Fig.~\ref{analysis} (a) and the semantic richness in Fig.~\ref{analysis} (b), the SSM exhibits a distinct and typical long-tail distribution. Only a tiny fraction of patches possess extremely high SSM values; while scarce in number, these patches are semantically “pure", allowing the model to anchor fundamental spatial-spectral mapping relationships by prioritizing the reconstruction of these ``elite" few.}

\subsection{Ablation Studies}
To verify the effectiveness of the proposed dynamic masking and knowledge injection strategies, we conduct ablation studies on SIGMAE by comparing it with the original MAE \citep{he2022masked}, fusion-based MAE that incorporates spectral indices as an additional modality \citep{bachmann2022multimae}, and a SIGMAE variant without random noise.
Compared with MAE and fusion-based MAE, both variants of SIGMAE achieve consistent improvements across all datasets, as shown in Table~\ref{tab:ablation}. Specifically, the introduction of the dynamic masking strategy $\gamma(e)$ leads to noticeable gains in mIoU and F1, indicating enhanced representation learning through adaptive feature reconstruction. The full version of SIGMAE, which integrates both dynamic masking and knowledge injection, achieves the highest performance in all metrics, demonstrating the effectiveness of the proposed strategies in improving model generalization and robustness across diverse remote sensing applications.

\begin{table*}[!htbp]
\centering
\caption{Ablation Studies (\%)}
\label{tab:ablation}

\setlength{\tabcolsep}{2.5pt} 
\renewcommand{\arraystretch}{1.2}
\footnotesize 
\newcolumntype{C}{>{\centering\arraybackslash}X}

\begin{tabularx}{\textwidth}{@{} l *{10}{C} @{}}
\toprule
Model Variants & \multicolumn{2}{c}{FOD} & \multicolumn{2}{c}{Wildfire} & \multicolumn{2}{c}{EuroSAT} & \multicolumn{2}{c}{SegMunich} & \multicolumn{2}{c}{OSCD} \\
% 使用 (lr) 截断横线，增加数据集之间的视觉间隙
\cmidrule(lr){2-3} \cmidrule(lr){4-5} \cmidrule(lr){6-7} \cmidrule(lr){8-9} \cmidrule(lr){10-11}
 & mIoU & F1 & mIoU & F1 & mIoU & mF1 & mIoU & mF1 & mIoU & F1 \\
\midrule
SIGMAE w/o SSDTM       & 59.19 & 65.96 & 89.92 & 89.96 & 94.54 & 97.15 & 46.16 & 59.98 & 65.04 & 74.51 \\
MAE w/ Index Fusion              & 59.70 & 67.03 & 89.06 & 89.04 & 95.16 & 97.50 & 46.31 & 59.96 & 64.96 & 74.44 \\
SIGMAE w/o $\gamma(e)$ & \textbf{61.23} & 68.78 & 90.67 & 90.73 & 94.58 & 97.18 & 46.44 & 60.16 & \textbf{66.07} & \textbf{75.70} \\
SIGMAE (Full)          & 61.21 & \textbf{68.87} & \textbf{91.12}& \textbf{91.20} & \textbf{95.53} & \textbf{97.70} & \textbf{47.45} & \textbf{60.90} & 65.45 & 75.03 \\
\bottomrule
\end{tabularx}
\end{table*}

\subsection{Reconstruction Performance}
\begin{figure*}
    \centering
\includegraphics[width=\linewidth]{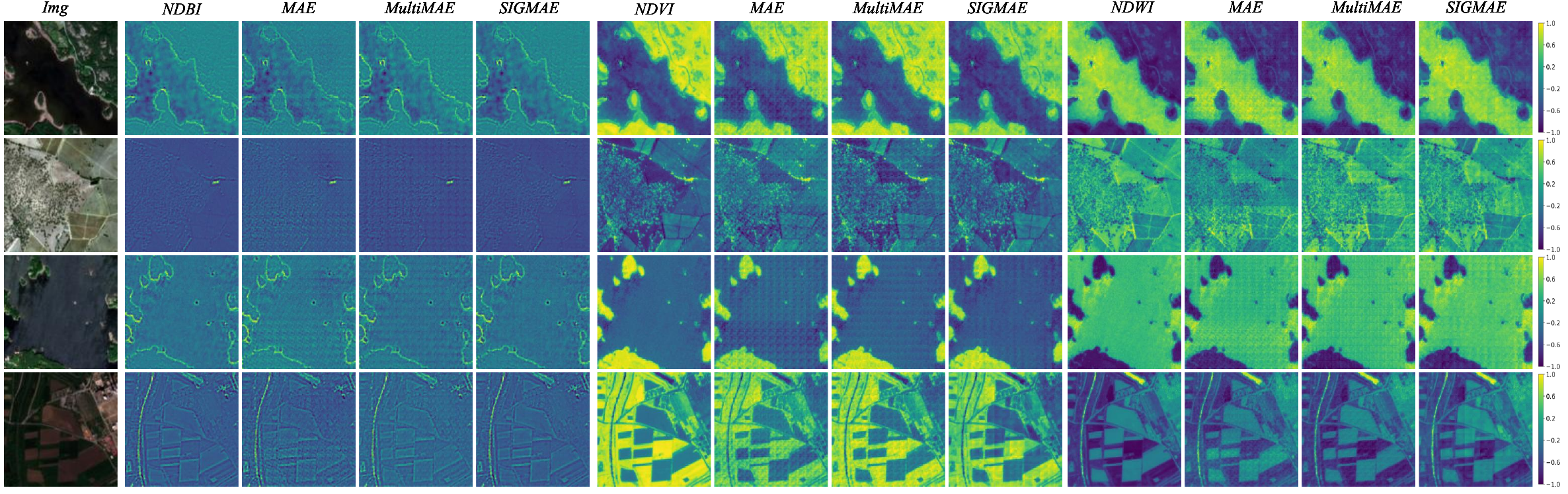}
    \caption{Spectral reconstruction performance comparison.}
    \label{fig:recon}
\end{figure*}

We have undertaken extensive investigations into the reconstruction performance of different models to validate the spectral modeling capability in the pretraining. From visual comparison of the reconstruction performance of different models, as shown in Fig.~\ref{fig:recon}, SIGMAE exhibits superior reconstruction fidelity compared to MAE and MultiMAE. 
The evident visual discrepancies observed in the reconstruction are primarily attributed to spectral degradation caused by the relatively limited reconstruction and inference capacities of the MAE model. The reconstructions of SIGMAE retain sharper details and exhibit fewer distortions, particularly in regions characterized by complex textures and transitions.

\subsection{{Masking Ratio Analysis}}
\begin{figure*}
    \centering
\includegraphics[width=\linewidth]{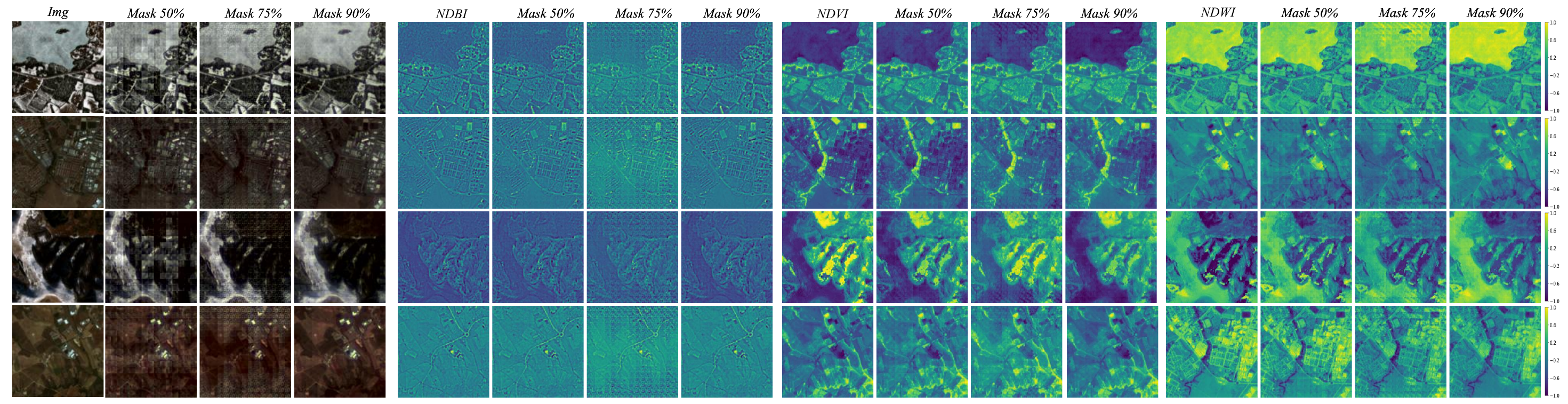}
    \caption{Image reconstruction performance comparison with varied masking ratios of $50\%$, $75\%$ and $90\%$, respectively.}
    \label{fig:mask_ratio}
\end{figure*}

We further provide visual comparisons across varying masking ratios ($50\%$, $75\%$, and $90\%$) to assess the spectral reconstruction performance of SIGMAE.
The capacity of the pre-training strategy to learn rich, generalized representations is visually confirmed by the reconstruction performance shown in Fig.~\ref{fig:mask_ratio}. 
As the masking ratio increases from 50\% to 90\%, the reconstructed images maintain high fidelity, preserving structural continuity, fine-grained textures, and the characteristic intensity distribution of the indices. This robust reconstruction capability, particularly across multiple spectral indices, demonstrates that the model successfully learns the underlying spatial and spectral dependencies within the remote sensing data.

A 90\% masking ratio leaves the model with very limited observable information, driving it to rely primarily on the global statistics or mean structure of the data. However, the model is still able to produce visually smooth and coherent reconstructions. However, the absence of sufficient contextual cues inevitably leads to the loss of fine-scale details. 
As a result, it converges more rapidly than models trained with lower masking ratios, as illustrated in Fig.~\ref{fig:loss}.
It shows that EuroSAT (classification) and SegMunich (large-area segmentation) gain from the high-ratio masking setting, whereas fine-grained datasets such as Wildfires and OSCD exhibit noticeably reduced performance.

\subsection{{Fine-Tuning Convergence}}
\begin{figure*}
    \centering
\includegraphics[width=\linewidth]{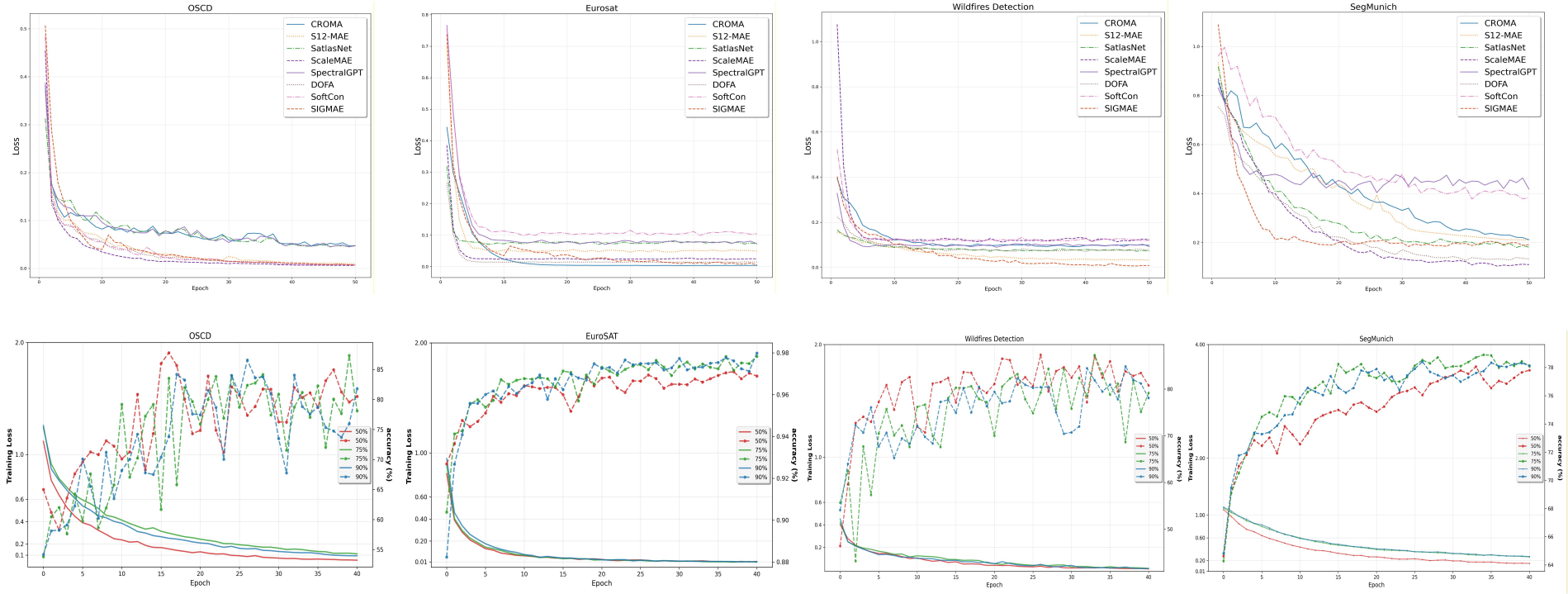}
    \caption{Convergence performance comparison with different models and different mask ratios.}
    \label{fig:loss}
\end{figure*}

The fine-tuning loss curves across four datasets offer key insights into model stability and convergence. As shown in Fig.~\ref{fig:loss}, the proposed SIGMAE achieves one of the fastest convergence rates, reducing loss rapidly within the first 10–15 epochs, especially on the EuroSAT and OSCD benchmarks. Moreover, SIGMAE maintains exceptional training stability, with smooth and consistent loss reduction and minimal oscillations, indicating robust optimization and effective regularization. In contrast, models such as SpectralGPT and SoftCon exhibit higher baseline losses and unstable convergence, with fluctuating or prematurely flattened loss curves (e.g., SoftCon on SegMunich). These results confirm that SIGMAE’s injection of prior knowledge supports stable, efficient learning and superior final optimization.

\subsection{Visual Comparison}
\begin{figure*}
    \centering
\includegraphics[width=\linewidth]{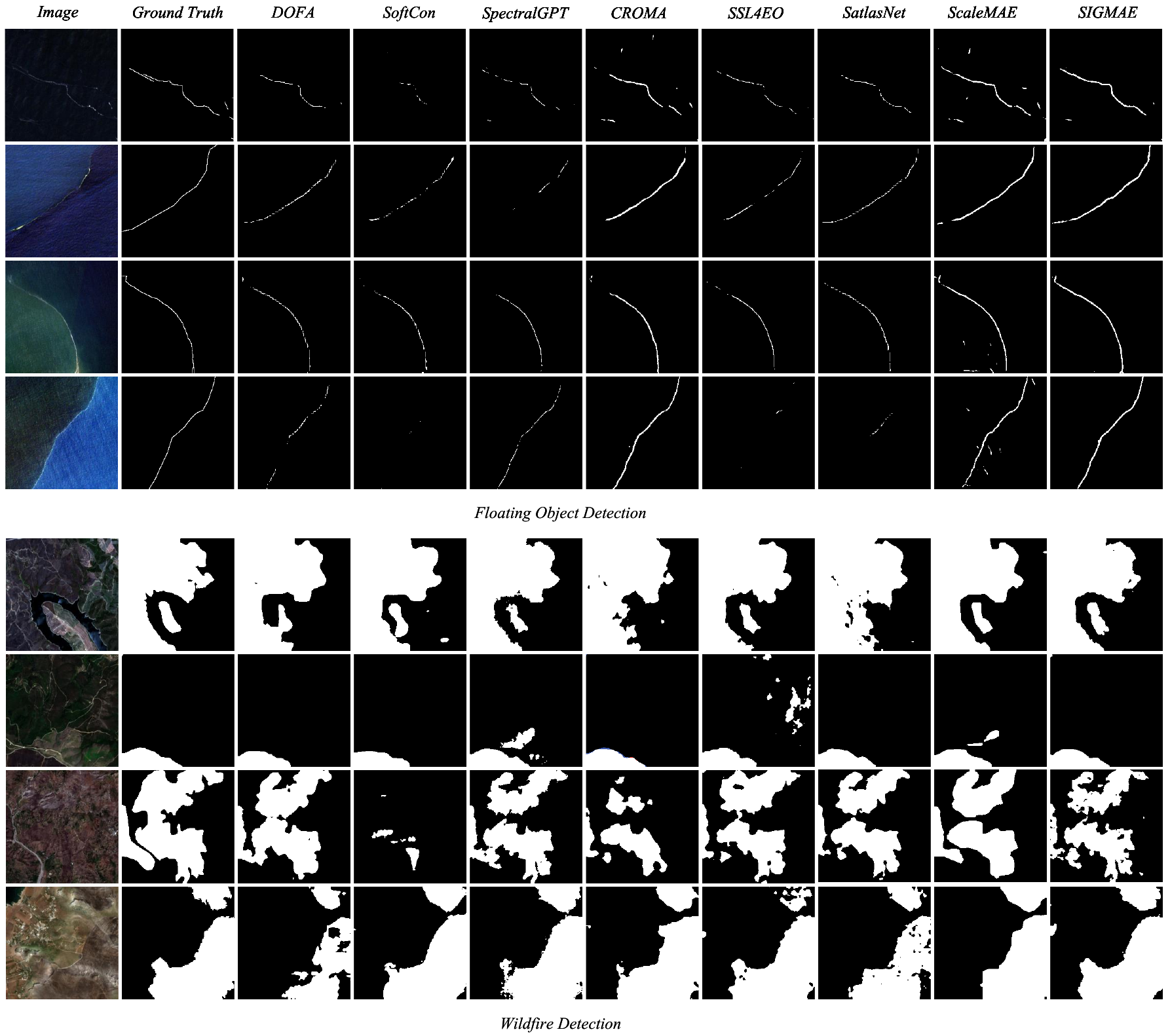}
\vspace{-9pt}
    \caption{Segmentation performance on the Floating object detection and wildfire detection datasets.}
    \label{fig:res_f}
\end{figure*}

\begin{figure*}
    \centering
\includegraphics[width=\linewidth]{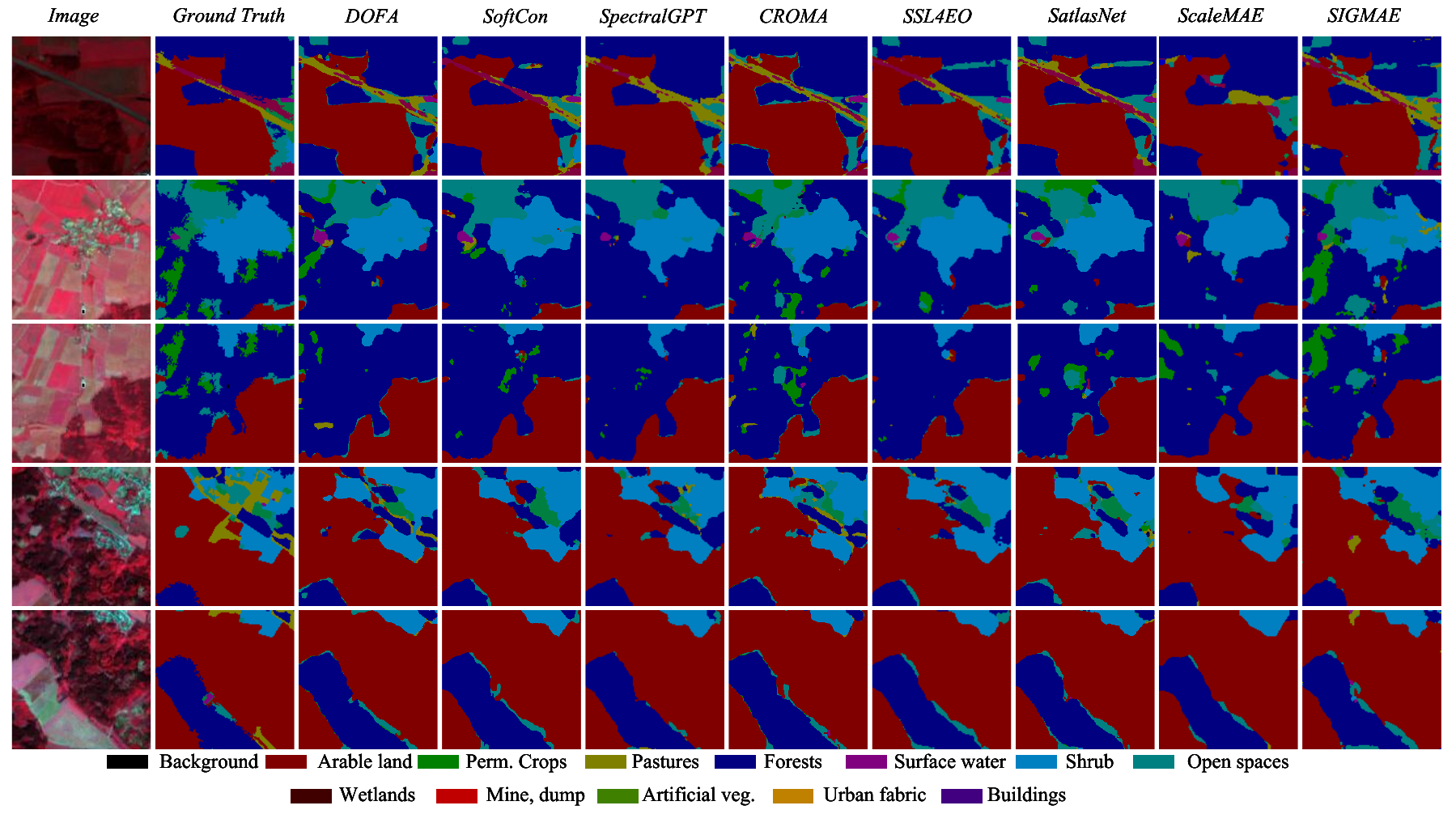}
\vspace{-9pt}
    \caption{Semantic segmentation performance on the SegMunich dataset.}
    \label{fig:segmunich}
\end{figure*}

\begin{figure*}
    \centering
\includegraphics[width=\linewidth]{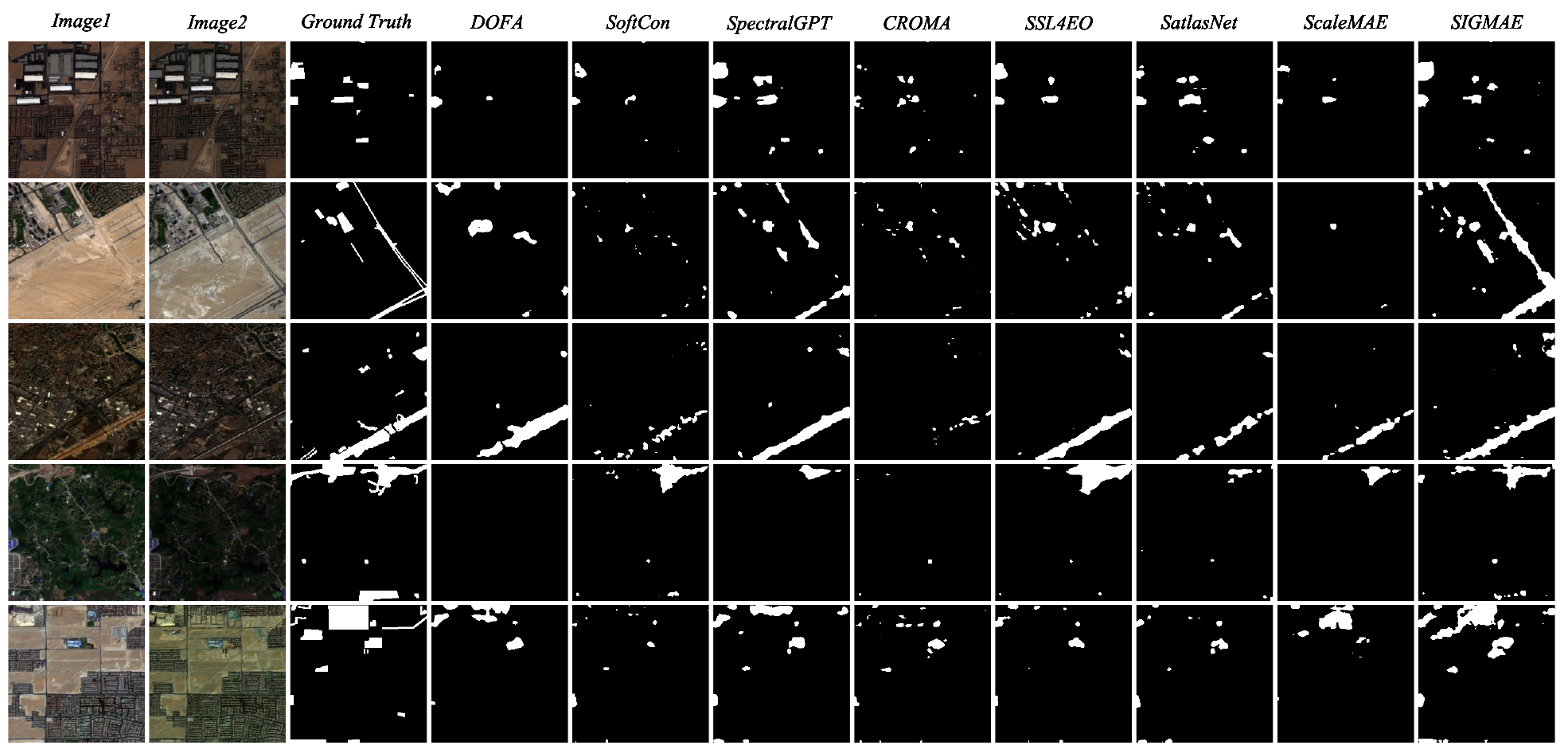}
\vspace{-9pt}
    \caption{Change detection performance on the OSCD dataset.}
    \label{fig:res_oscd}
\end{figure*}

The visual comparison of segmentation and classification results is displayed in Figs.~\ref{fig:res_f},~\ref{fig:segmunich} and ~\ref{fig:res_oscd}. In Floating Object Detection, where targets are primarily linear, competing methods such as SpectralGPT and SoftCon often produce fragmented masks, resulting in false negatives and compromised structure. SIGMAE, in contrast, generates continuous, high-precision boundaries closely matching the ground truth. For Wildfire Detection, involving complex, large-area wildfire areas, SIGMAE robustly delineates irregular boundaries with fewer internal voids or under-segmentation than DOFA and CROMA, highlighting its strong generalization and noise suppression capabilities.

In semantic segmentation on the SegMunich dataset, many methods struggle with fine-grained boundaries, resulting in jagged edges and misclassified small structures. SpectralGPT and AtlasNet produce blurred transitions and blocky artifacts between major classes such as ``Forests'' and ``Arable land''. SIGMAE delivers precise, coherent masks while preserving small land cover patches like thin roads or fragmented ``Permanent Crops'', leveraging both spectral and contextual information for accurate pixel-level classification.

For change detection on the OSCD dataset, which features small, isolated changes and spectral noise from atmospheric effects, DOFA, SoftCon, and ScaleMAE generate numerous false positives in background areas. SIGMAE mitigates this issue via its dynamic masking mechanism, producing clean change maps while maintaining sharp boundaries for small changes such as new buildings, achieving a superior balance between noise suppression and recall. Overall, SIGMAE consistently provides accurate and structurally coherent segmentation and detection across varied tasks and challenging conditions.

\subsection{Scope and Limitations}
The proposed SIGMAE focuses on pretraining of large-scale multispectral images by exploiting remote sensing index information to guide the masking process. Compared to the multimodal visual foundation \citep{guo2024skysense}, the limitation of SIGMAE lies in that it is specially designed for a single modality, i.e., multispectral images. Our goal is to explore an efficient pretraining and fine-tuning strategy on multispectral images.
And the model performance inherently depends on the domain knowledge, namely, the spectral indices. For downstream tasks, we evaluate the performance of our method on several mainstream multispectral datasets. In future work, we plan to incorporate multimodal datasets for pre-training, which is expected to further enhance the robustness and generalization capability of SIGMAE across diverse Earth observation applications.

\section{Conclusion}\label{sec5}
This paper proposed a novel spectral index-guided Masked Autoencoder (SIGMAE) to build a foundation model for multispectral remote sensing images.
It deploys MAE-style self-supervised pretraining, followed by fine-tuning with a limited number of labeled samples. 
Meaningful spectral indices are incorporated as prior domain knowledge to enhance the feature representation of the target objects. Instead of directly reconstructing spectral indices or using them as additional model inputs, we propose a novel dynamic masking strategy that leverages spectral indices to replace the random sampling strategy used in the original MAE.
By guiding the masking process towards informative regions during image reconstruction, this strategy improves the model’s ability to capture essential object properties, ensuring that masked regions contribute to a more meaningful and informative learning process.
Experiments on five large-scale datasets validate the effectiveness of the proposed approach. The results show that the proposed SIGMAE outperforms existing MAE-based methods in both spatial and spectral information reconstruction. More importantly, it learns more robust feature representations, leading to improved classification, segmentation and change detection performance. 
Future work will focus on extending the proposed dynamic masking strategy to multimodal remote sensing image pretraining, utilizing various domain-specific knowledge.

% \subsection*{Author Contributions} 
% X. Zhang conceived the idea and designed the experiments.  
% B. Li and C. Zhou conducted the experiments and analyzed the data.  
% W. Yu contributed to the interpretation of the results and manuscript writing.  
% All authors contributed equally to the writing of the manuscript.

% \section*{Funding}
% % \subsection*{Funding} 
% This research was financially supported by the Open Research Fund from Guangdong Laboratory of Artificial Intelligence and Digital Economy (SZ), under Grant No. GML-KF-24-28, and was also supported in part by the National Natural Science Foundation of China under Grant No. 42371374.

% --- 8. 文末声明部分 (对应样刊 image_9ebd43.png 效果) ---

% 作者贡献声明
\printcredits

% 利益冲突声明
\section*{Declaration of competing interest}
The authors declare that they have no known competing financial interests or personal relationships that could have appeared to influence the work reported in this paper.

% 致谢与基金 (资助信息放在这里)
\section*{Acknowledgments}
This work was supported in part by the Open Research Fund from Guangdong Laboratory of Artificial Intelligence and Digital Economy (SZ) under Grant No. GML-KF-24-28, and in part by the National Natural Science Foundation of China under Grant No. 42371374.

% --- 9. 参考文献 ---
\bibliographystyle{cas-model2-names}
\bibliography{cas-refs}

\end{document}